\documentclass[letterpaper, 10 pt, conference]{ieeeconf}  

\IEEEoverridecommandlockouts                              

\usepackage{graphics} 
\usepackage{epsfig} 
\usepackage{mathptmx} 
\usepackage{times} 
\usepackage{amsmath} 
\usepackage{amssymb}  

\title{\LARGE \bf
Encoding Cardiopulmonary Exercise Testing Time Series as Images for Classification using Convolutional Neural Network
}

\author{Yash Sharma$^{1}$, Nick Coronato$^{2}$ and Donald E. Brown$^{2}$
\thanks{$^{1}$Department of Pediatrics, University of Virginia
        {\tt\small ys5hd@virginia.edu}}%
\thanks{$^{2}$Department of Engineering Systems and Environment, University of Virginia}%
}

\begin{document}

\maketitle
\thispagestyle{empty}
\pagestyle{empty}

\begin{abstract}

Exercise testing has been available for more than a half-century and is a remarkably versatile tool for diagnostic and prognostic information of patients for a range of diseases, especially cardiovascular and pulmonary. With rapid advancements in technology, wearables, and learning algorithm in the last decade, its scope has evolved.
Specifically, Cardiopulmonary exercise testing (CPX) is one of the most commonly used laboratory tests for objective evaluation of exercise capacity and performance levels in patients. CPX provides a non-invasive, integrative assessment of the pulmonary, cardiovascular, and skeletal muscle systems involving the measurement of gas exchanges. However, its assessment is challenging, requiring the individual to process multiple time series data points, leading to simplification to peak values and slopes. But this simplification can discard the valuable trend information present in these time series. In this work, we encode the time series as images using the Gramian Angular Field and Markov Transition Field and use it with a convolutional neural network and attention pooling approach for the classification of heart failure and metabolic syndrome patients. Using GradCAMs, we highlight the discriminative features identified by the model. 
\newline

\indent \textit{Clinical relevance} — The proposed framework can process multivariate exercise testing time-series data and accurately predict cardiovascular diseases. Interpretable Grad-CAMs can be obtained to explain the prediction.  
\end{abstract}

\section{INTRODUCTION}

Exercise intolerance is a major clinical feature from the early stages and a source of symptoms for referral to a physician. In the last decade, Cardiopulmonary Exercise Testing (CPX) has emerged as an important tool for non-invasive monitoring of cardiopulmonary vitals of patients. It provides an objective, reliable, and reproducible assessment of cardiorespiratory fitness and, as such, an effective instrument for use by clinical practitioners to inform strategies to improve the health outcomes of their patients. CPX enables the measurement of physiological response to physical exercise through an array of pulmonary, cardiovascular, and metabolic measurements built around breath-by-breath gas exchange analysis. In cardiology, CPX was introduced in early 1980s for the classification of patients with Heart failure with reduced ejection fraction 
\cite{weber1982oxygen}. As physical work is an essential activity of daily living, CPX is “disease agnostic” and, over the years, has been adopted as an assessment tool for a wide range of conditions such as heart failure, hypertension, Gaucher’s disease, and cystic fibrosis, among others \cite{guazzi2017cardiopulmonary, guazzi2007exercise, bjelobrk2021cardiopulmonary, williams2019cardiopulmonary}. However, the very integrative and complex physiology revealed by CPX renders analysis challenging–its interpretation requires the individual to process a bewildering array of thousands of data points, including oxygen uptake, carbon dioxide output, ventilation, heart rate, all of which are changing dynamically with time as the exercise performed starts, stops, and alters in intensity.

In 2010, in its scientific statement, AHA highlighted that CPX provides a wide array of unique and clinically useful incremental information that has been poorly understood and underutilized by practicing physicians \cite{balady2010clinician}. Interpreting multiple time series variables obtained in CPX is a non-trivial time-consuming exercise. The interpretation requires extensive knowledge and a detailed understanding of all variables, tables, and flow charts. However, recent growth in machine learning (ML) research has propelled the applications of CPX. Inbar et al. \cite{inbar2021machine} used SVM for identifying chronic heart failure patients using CPX, Yang et al. \cite{yang2019use} utilized ML for studying the efficacy of aerobic exercise intervention on young hypertensive patients, Diller et al. \cite{diller2019machine} used deep learning on CPX variables along with other clinical, demographic, ECG, and laboratory parameters for guiding therapy in adult congenital heart disease. Moreover, due to strenuous testing involved in CPX, active research is being conducted to develop ML-based approaches for early estimation of patient response to exercise \cite{baralis2015predicting} and for estimating the VO2 dynamics from the heart rate and inputs from the treadmill ergometer, cycle, and accelerometer, with focus on incorporating these algorithms into smart devices \cite{shandhi2020wearable, beltrame2016estimating, mazzoleni2017multi, akay2017development, de2018data, borror2019predicting}.

Interpretation of CPX is a multivariate time series problem involving simultaneous assessment of generated heart rate, ventilation, gas exchange (oxygen uptake), and carbon dioxide output. The manual evaluation and traditional analytics of these time series are simplified to peak values, summary indices, and slopes. VO2 peak is considered the gold standard assessment of cardiorespiratory fitness. This process of simplification and distilling signals out of long time series is known as feature engineering in ML. However, this process is constrained by the expertise of users and already discovered metrics in the literature, which can lead to the discarding of valuable time-series information. In recent years, Deep learning (DL) has emerged as the new flexible learning framework in which automatic extraction of relevant features happens based on the learning objective. Myers et al. \cite{myers2014neural} demonstrated that ANN performed better than conventional survival analysis on CPX data for estimating cardiovascular mortality risk. 

Among these multiple DL frameworks, due to parameter efficiency and spatial invariance property, CNN has gained adoption for many problems, including time series \cite{ciaburro2021machine}. Further, Wang et al. \cite{wang2015encoding} and Yang et al. \cite{yang2020sensor} demonstrated that encoding time series as images using Gramian Angular Field (GAF) and Markov Transition Fields (MTF) for CNNs leads to superior performance and in the identification of patterns not found in the one-dimensional sequence methods. The image encoding step in combination with small CNN leads to a parameter efficient framework, reducing the requirement of a big dataset required for training parameter-heavy DL models. In this paper, we used GAF/MTF for encoding CPX variables and used the attention pooling approach for aggregating the multivariate time series representation. Further, we demonstrated on the open Wafer data \cite{olszewski2001generalized} that the attention pooling approach leads to superior performance than the concatenation approach \cite{ilse2018attention}.

\begin{figure}[t]
\centering
\centerline{\includegraphics[width=\linewidth]{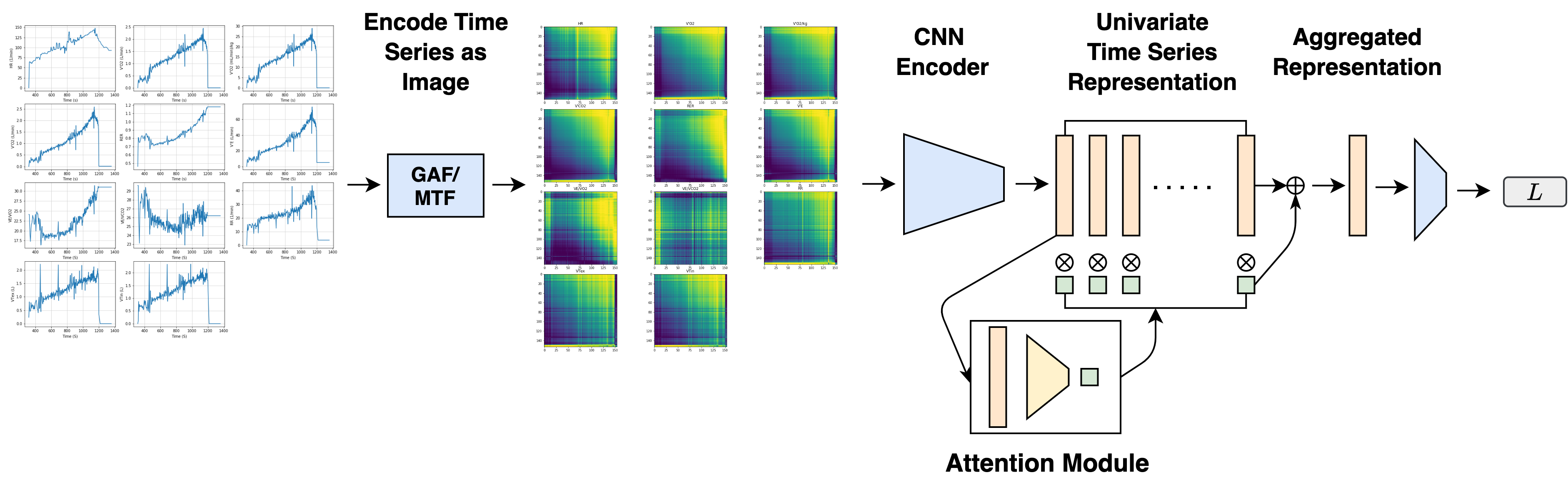}}
\caption{Proposed Architecture: Each of the univariate time series is encoded as an image using GAF/MTF, passed through CNN model for generating representation, aggregated using attention pooling approach to a single representation, and passed through a linear layer for classification loss.}
\label{fig:gaf_arch}
\end{figure}

\section{Methods}

\subsection{Gramian Angular Field}

GAF encodes a time series into images by polar coordinates based matrix. First, a Piecewise aggregate approximation is performed to reduce the length of time series for aligning them to same length then values are normalized between -1 and 1. Rescaled values are encoded as angular cosine and time stamp as radius for encoding the values into polar coordinates. Encoded values warp among different angular points on the spanning circles with an increase in time. The encoding is bijective, and as opposed to cartesian coordinates, preserves absolute temporal relations. These rescaled time series can be used in 2 different forms: Summation Field (GASF) and Difference Field (GADF). These formulations preserve the temporal dependency since time increases as the position moves from top-left to bottom-right and embeds the temporal correlation within different time intervals in the image. 




\subsection{Markov Transition Field}

Markov Transition field uses Markov transition probabilities to preserve the information in the time domain. A $Q \times Q$ Markov transition matrix is created by dividing the data into Q quantile bins. The quantile bins that contain the data at time stamps $i$ and $j$ are $q_i$ and $q_j$. $M_{ij}$ in MTF denotes the transition probability of $q_i \rightarrow q_j$. The MTF encodes the multi-span transition probabilities of the time series. The MTF size is reduced by averaging the pixels in each non-overlapping $m \times m$ patch. However, as the probabilities of moving elements form the transformed matrix, the MTF method cannot revert to the raw time-series data like GAF, and it is not as symmetrical as the GAF method. 



\begin{figure}[t]
\centering
\includegraphics[width=\linewidth]{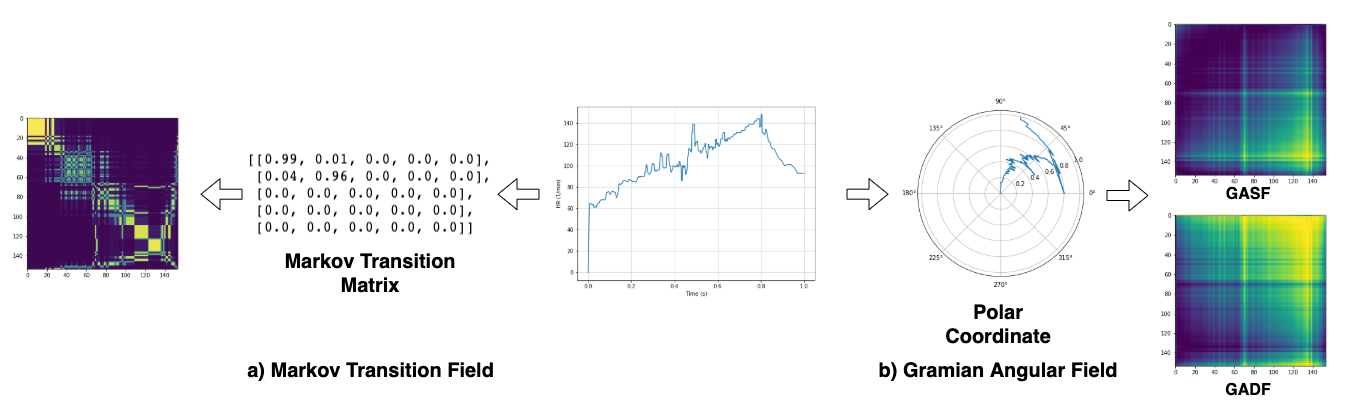}
\caption{Illustration of encoding methods: a) Markov Transition Field b) Gramian Angular Field}
\label{fig:polar}
\end{figure}

\subsection{Attention Pooling}
		
We used the weighted-average aggregation for aggregating the univariate time series representation to multivariate time series representation. This pooling approach uses a two-layered neural network to compute weights for each time series and use them for aggregating representation \cite{ilse2018attention}.


\subsection{Set-Up}

The model was implemented with PyTorch and trained on a single RTX2080 GPU. The framework was trained end-to-end with Adam optimizer with a batch size of 1 and for 100 epochs. Based on the validation accuracy, the best-performing model was picked. A learning rate of 0.0023 was used in Wafer experiments and 3e-4 in CPX experiments. 

\section{Results}

\subsection{Data Description}

We demonstrated the approach on open-sourced Wafer data for comparison to previous approaches and reported our results on CPX data. The CPX dataset consists of 30 patients diagnosed with either heart failure or metabolic syndrome (15 patients each). For each of the patients, CPX data contain breath-by-breath readings of the following variables: Metabolic equivalent of task (1 MET = 3.5ml/kg/min); HR (beats/min); Absolute VO2 (L/min); Relative VO2 adjusted to body mass (ml/kg/min); VCO2 (L/min); Respiratory exchange ratio; VE (L/min); VE/VCO2; VE/VO2; respiratory rate (breaths/min); expiratory tidal volume (L); and inspiratory tidal volume (L). Proper ethical review was obtained for using this data in the study. 
The Wafer dataset was collected from six vacuum chamber sensors that monitored the manufacture of semiconductor microelectronics and have two classes normal and abnormal \cite{olszewski2001generalized}. For experimentation, five-fold cross-validation was used, and results averaged from 20 runs were reported. 

\subsection{Wafer Data}

\begin{figure}[t]
\centering
\includegraphics[width=\linewidth]{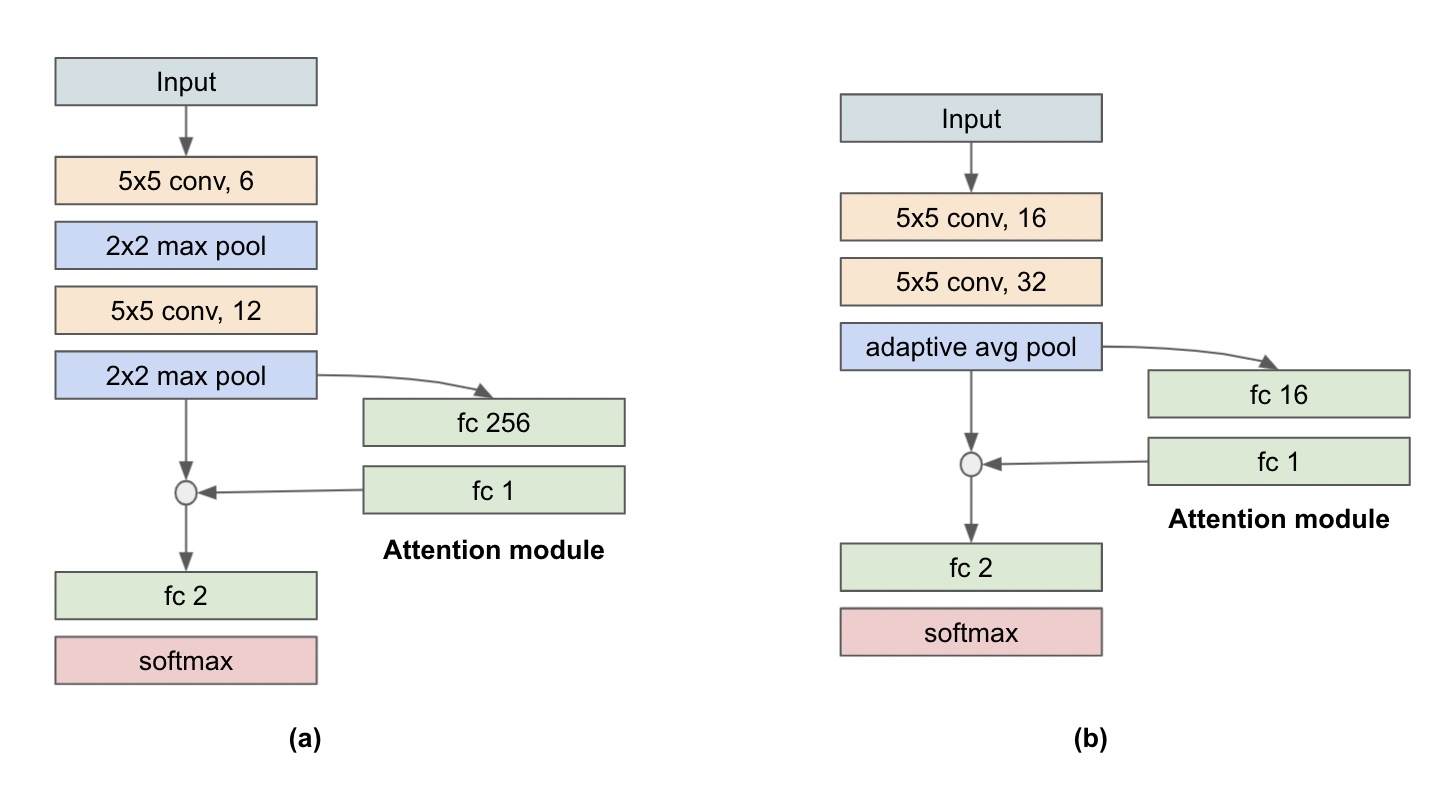}
\caption{CNN Architecture: a) Proposed in \cite{palm2012prediction} and used for Wafer Data Experiment b) Used in CPX data experiments}
\label{fig:cnn_arch}
\end{figure}

We compared our approach to other proposed approaches for Wafer data \cite{olszewski2001generalized, yang2020sensor}. For consistency, we use the same encoder architecture used in \cite{yang2020sensor}, originally proposed in \cite{palm2012prediction} (Figure~\ref{fig:cnn_arch}). We demonstrated that attention pooling leads to lower error rates than the concatenation approach. Moreover, we highlight that CNN-based methods on image encoded time series perform competitively to the other approaches. Based on our observations, we hypothesize that for multivariate time series problems, a small CNN model on GAF/MTF based encoding of time series with attention pooling is capable of strong baselines in a limited data scenario.

\begin{table}[t]
  \label{model-comparison}
  \caption{Comparison of average classification error rates (\%) of different methods on the Wafer dataset and performance of our proposed approach on CPX dataset.}
  \centering
  \begin{tabular}{ccc}
    \hline
    \textbf{Approach}     & \multicolumn{2}{c}{\textbf{Error (\%)}}\\
     & \textbf{Wafer} & \textbf{CPX} \\
    \hline
    DTW \cite{gorecki2015multivariate}   & 2.01 & - \\
    DDTW \cite{gorecki2015multivariate}      & 9.21 & - \\
    DDDTW \cite{gorecki2015multivariate}     & 1.92 & - \\
    STKG-SVM-K3 \cite{prieto2015stacking}     & 1.23 & - \\
    STKG-NB-K5 \cite{prieto2015stacking}     & 3.69 & -    \\    
    STKG-IF-PSVM-DT+M \cite{prieto2015stacking}     & 0.84 & -    \\
    STKG-IF-NB-SVM+M \cite{prieto2015stacking}     & 2.23 & -   \\    
    normDTW \cite{luczak2018combining}     & 3.85 & -    \\
    combDTW \cite{luczak2018combining}     & 2.01 & -    \\    
    LSTM-FCN \cite{karim2019multivariate}     & 1.00 & -    \\    
    MLSTM-FCN \cite{karim2019multivariate}     & 1.00 & -  \\        
    ALSTM-FCN \cite{karim2019multivariate}     & 1.00 & -   \\        
    MALSTM-FCN \cite{karim2019multivariate}     & 1.00 & -  \\        
    MALSTM-FCN \cite{karim2019multivariate}     & 1.00 & -  \\      
    concat-MTF-RGB \cite{yang2020sensor}     & 0.40 & -\\
    concat-GASF-RGB \cite{yang2020sensor}     & 0.57 & -\\
    concat-GADF-RGB \cite{yang2020sensor}     & 0.44 & -\\
    concat-MTF (ours)   & 0.63  & 23.40 \\
    concat-GASF (ours)   & 1.18 & 12.34 \\
    concat-GADF (ours)   &  0.46 & 8.67 \\
    Attn-MTF (ours)   & 0.41 & 15.00 \\
    Attn-GASF (ours)  & 0.55 & 16.17 \\
    Attn-GADF (ours)  & \textbf{0.15} & \textbf{7.67} \\
    \hline
  \end{tabular}
\end{table}

\subsection{CPX Data}
\label{cpx_interpretation}

\begin{figure}[b]
\centering
\includegraphics[width=\linewidth]{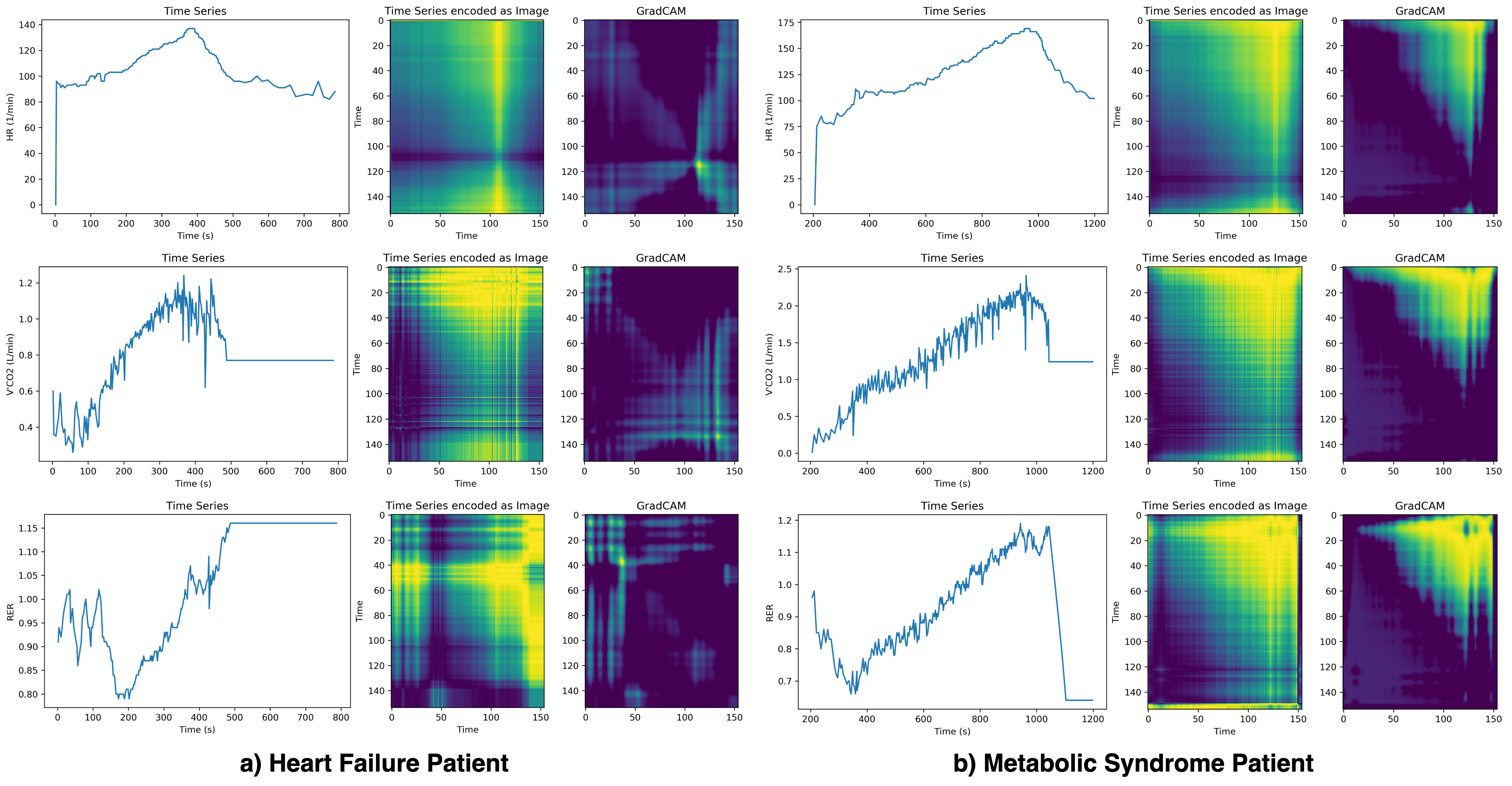}
\caption{GradCAM examples of the top attended patches (HR, RER, V'CO2). The part highlighted as green was considered relevant by the models. Interpretation of charts is discussed in Section ~\ref{cpx_interpretation}}
\label{fig:gaf_cam}
\end{figure}

In the CPX data, for high-quality GradCAMs, we made minor changes to the CNN architecture used in Wafer experiments, removed the max-pooling layer and replaced it with single adaptive average pooling, and increased the number of channels in both the layers (Figure~\ref{fig:cnn_arch}). Among the three time-series encoding approaches - GASF, GADF, and MTF, GADF was performing best. We attributed the high performance to the temporal changes that GADF captured during the CPX exercise. In the GADF approach, HR, RER, and V'CO2 were assigned the highest attention among all the time series. We analyzed these frames using GradCAMs to probe the patterns CNN identified for differentiating diseases. In metabolic syndrome patients, we observed that CNNs focused on the increase in the slope of the time series. For heart failure patients, we observed that the model was sensitive to alterations happening in quick succession and to the drop in time series during exercise (Figure~\ref{fig:gaf_cam}). Since all the images were encoded independently using GAF/MTF methods, the model was exposed to temporal trends instead of absolute values, which identified trends unique to heart failure and metabolic syndrome. The patterns identified in the GradCAM plots were deemed relevant in our medical review by a panel of cardiologists and pulmonologists.  

\section{Conclusions}
\label{conc}

In this paper, we demonstrated the strong performance of small CNNs on CPX data for differentiating heart failure and metabolic syndrome patients. With the advancement in the internet of things and wearables, a large amount of fitness data is getting collected, which can be used with deep learning modeling to diagnose diseases. Similar approaches like ours can aid experts in identifying markers relevant for the early diagnosis of health conditions. Our approach is able to accurately classify heart failure and metabolic syndrome patients using temporal trends instead of simplified values such as peak and slope. Further, we highlighted that the attention pooling approach could be used with GAF/MTF approaches to aggregate univariate time series representation. 
Interpretation based on this approach has vast scope in the field of exercise testing.



\section*{ACKNOWLEDGMENT}

This work was supported by the National Center for Advancing Translational Science of the National Institutes of Health Award UL1TR003015/ KL2TR003016.

We would like to thank Dr. Arthur L. Weltman and Exercise Physiology Core Laboratory for providing the data and helping us with queries.


\bibliographystyle{IEEEtran}
\bibliography{IEEEabrv}

\end{document}